\pdfoutput=1

\documentclass[11pt]{article}

\usepackage{acl}

\usepackage{times}
\usepackage{latexsym}

\usepackage[T1]{fontenc}

\usepackage[utf8]{inputenc}

\usepackage{microtype}
\usepackage{algorithm}
\usepackage{algorithmic}
\usepackage{enumitem}
\usepackage{multirow}
\usepackage{graphics}
\usepackage{booktabs}
\usepackage{newfloat}
\usepackage{listings}
\usepackage{amsmath,bm}
\usepackage{amsthm}
\usepackage{float}
\usepackage{makecell}

\usepackage{pdfpages}
\usepackage{titlesec}
\usepackage{subcaption}
\usepackage{enumitem}
\setenumerate[1]{itemsep=0pt,partopsep=0pt,parsep=\parskip,topsep=5pt}
\setitemize[1]{itemsep=0pt,partopsep=0pt,parsep=\parskip,topsep=5pt}
\usepackage{wasysym}
\usepackage{upgreek}
\usepackage{color}

\title{Improving Meta-learning for Low-resource \\Text Classification and Generation via Memory Imitation}

\author{Yingxiu Zhao$^{1}$, \ \ Zhiliang Tian$^{1}$\thanks{\ \ Corresponding author}, \ \ Huaxiu Yao$^{2}$, Yinhe Zheng$^{3}$, Dongkyu Lee$^{1}$\\
\textbf{Yiping Song$^{4}$,\ \ Jian Sun$^{3}$,\ \ Nevin L. Zhang$^{1}$} \\
\normalsize{$^1$The Hong Kong University of Science and Technology, Hong Kong SAR, China}\\
\normalsize{$^2$Stanford University}, 
\normalsize{$^3$Alibaba Group}\\
\normalsize{$^4$Department of Computer Science, Peking University, Beijing, China}\\
{\small\tt \{yzhaocx,ztianac,dleear,lzhang\}@connect.ust.hk,huaxiu@cs.stanford.edu}\\
{\small\tt \{zhengyinhe.zyh,jian.sun\}@alibaba-inc.com,songyiping@pku.edu.cn},
{\small\tt } \\
}

\begin{document}
\maketitle
\begin{abstract}
Building models of natural language processing (NLP) is challenging in low-resource scenarios where only limited data are available.
Optimization-based meta-learning algorithms achieve promising results in low-resource scenarios by adapting a well-generalized model initialization to handle new tasks.
Nonetheless, these approaches suffer from the \textit{memorization overfitting} issue, 
where the model tends to memorize the meta-training tasks while ignoring support sets when adapting to new tasks.
To address this issue, we propose a memory imitation meta-learning (MemIML) method that enhances the model's reliance on support sets for task adaptation. 
Specifically, we introduce a task-specific memory module to store support set information and construct an imitation module to force query sets to imitate the behaviors of some representative support-set samples stored in the memory.
A theoretical analysis is provided to prove the effectiveness of our method, and empirical results also demonstrate that our method outperforms competitive baselines on both text classification and generation tasks.
\end{abstract}

\section{Introduction}
Building natural language processing (NLP) models in low-resource scenarios is of great importance in practical applications because labeled data are scarce.
Meta-learning-based methods \citep{thrun2012learning} have been commonly used in such scenarios owing to their fast adaptation ability.
Notable successes have been achieved by meta-learning on low-resource NLP tasks, such as multi-domain sentiment classification \citep{arsc,induction} and personalized dialogue generation \citep{paml,song2019learning,zheng2020pre}. 

Among different meta-learning approaches \citep{9428530}, optimization-based approaches have been widely used in various low-resource NLP scenarios \citep{paml,qian2019domain,metamt,mi2019meta} because they are model-agnostic and easily applicable. Concretely, optimization-based meta-learning algorithms aim to learn a well-generalized global model initialization $\theta$ that can quickly adapt to new tasks within a few steps of gradient updates.
In the meta-training process, we first train $\theta$ on a \textit{support set} (i.e., \ a few training samples of a new task $i$) to obtain task-specific parameters $\theta'_i$. Then, we optimize $\theta$ based on the performance of $\theta'_i$ on a \textit{query set} (i.e., another set of samples in task $i$).

Despite its effectiveness, optimization-based meta-learning algorithms usually suffer from the \textit{memorization overfitting} issue
\footnote{Memorization overfitting is different from the overfitting in conventional supervised learning \cite{tra_overfit}. The latter means that the model overﬁts to the training tasks and fails to generalize to the testing tasks.}
\citep{mm_overfit,meta_aug},
where the learned model tends to solve all the meta-training tasks by memorization, rather than learning how to quickly adapt from one task to another via support sets. 
This is acceptable for training process, but results in poor generalization on the meta-testing sets, because the memorized model does not have knowledge of those tasks and does not know how to utilize the base learner to learn new tasks.
Hence, this issue hinders the model from capturing task-specific characteristics from support sets and thus prevents the model from adapting to distinct new tasks \citep{meta_aug}.
For instance, in personalized dialogue generation, this implies that the dialog model cannot adapt to individual users based on short conversation histories and hence fails to generate personalized responses.

Several works have been proposed to tackle the memorization overfitting issue for regression and image classification tasks.
Some studies try to explicitly regularize the model parameters \citep{mm_overfit,meta_aug}, but this restricts the complexity of model initialization and reduces the model capacity.
Another line of research integrates samples from support sets into the corresponding query sets via data augmentation \citep{task_aug}. 
However, data augmentation on textual data may result in noisy labels or distribution shifts, which impairs the model performance \citep{chen2021empirical}. 

In this paper, we address the memorization overfitting issue by enhancing the model's dependence on support sets when learning the model initialization, which forces the model to better leverage information from support sets. 
As an analogy, 
consider a young investor who has the ability to adapt to new circumstances rapidly but little memory of learned experiences, and an old investor who is experienced but refuses to be flexible.
Our idea is to make the young investor adaptive to the various situations when he assesses his benefits so that he can not only take advantage of the old one's experience but also learn from the old investor how to leverage the learned experience.
In this paper, the young investor stands for a standard meta-learning algorithm (e.g., MAML), which is prone to memorization overfitting, and the old investor is a memory module we integrate into the method, carrying information of support sets.

Specifically, we propose a \underline{Mem}ory-\underline{I}mitation \underline{M}eta-\underline{L}earning (MemIML) method that
forces query set predictions to depend on their corresponding support sets by dynamically imitating behaviors of the latter.
We therefore, introduce a memory module and an imitation module to enhance such dependence.
The memory module is task-specific, storing representative information of support sets. 
The imitation module assists in predicting samples of query sets by dynamically imitating the memory construction.
In this way, the model has to access the support set by memory imitation each time it makes a prediction on a query-set sample, hence it's no longer feasible for the model to memorize all meta tasks.

The contributions of this work are:
\begin{enumerate}[leftmargin=*]
\item A novel method MemIML is proposed to alleviate the memorization overfitting for optimization-based meta-learning algorithms. It encourages the utilization of support sets with the help of a memory module and an imitation module when adapting to new tasks.
\item Comprehensive experiments on text classification and generation tasks show that MemIML significantly outperforms competitive baselines.
\item Theoretical proofs are given to demonstrate the effectiveness of our method.
\end{enumerate}
\section{Related Work}
\paragraph{Meta-Learning.}
Meta-Learning aims to improve the learning algorithm itself based on the previously learned experience \citep{1998Learning,9428530}. In general, there are three categories of meta-learning methods: model-based methods, \citep{santoro2016meta,obamuyide2019meta} which depend on the particular model design to facilitate fast learning; metric-based methods, \citep{vinyals2016matching,snell2017prototypical,induction} which encode samples into an embedding space and classify them based on the learned distance metric; optimization-based methods \citep{maml,mi2019meta} that learn a well-generalized model initialization which allows for fast adaptation to new tasks. 
For low-resource scenarios in NLP, optimization-based meta-learning methods achieved promising results on tasks such as personalized dialog generation \citep{paml,song2019learning,tian2021learning}, low-resource machine translation \citep{gunmt,sharaf2020meta} and question answering \citep{yan-etal-2020-multi-source}, few-shot slot tagging \citep{mcml}, and so on.

\paragraph{Memorization overfitting of Meta-learning.}
Meta-learning algorithms suffer from memorization overfitting.
\citet{mm_overfit} build an information bottleneck to the model, while this approach decreases the model performance with this passive regularization.
\citet{meta_aug} inject random noise to the ground truth of both support and query sets, while little extra knowledge is introduced to learn a good initialization.
\citet{task_aug} address overfitting issues by augmenting meta-training tasks through mixing up support and query sets. However, such augmentation for text needs to be based on the assumption of keeping the label and the data distribution unchanged, which is often not true in practice \citep{chen2021empirical}.
Instead of regularization and data augmentation, we leverage the support sets information stored in the memory to augment the meta-learning.

\paragraph{External Memory for Few-shot Learning.} Memory mechanism has proven to be powerful for few-shot learning \citep{induction,santoro2016meta,metalearned}. 
Current methods either refine representations stored in the memory \citep{ramalho2018adaptive} or refining parameters using the memory \citep{munkhdalai2017meta,cai2018memory,fewnlp}. In the NLP domain, some methods store encoded contextual information into a memory \citep{kaiser2017learning,holla2020learning,zheng2019personalized}.
\citet{induction} propose a memory induction module with a dynamic routing algorithm for few-shot text classification tasks.
\citet{metalearned} augment the model with an external memory by learning a neural memory.
\citet{mcml} reuse learned features stored in the memory on the few-shot slot tagging.

\section{Preliminaries}
We first formulate model-agnostic meta-learning (MAML) \citep{maml}.
Specifically, denote the base model used in MAML as $f_\theta$ and assume each task $\mathcal{T}_i$ sampled from a task distribution $p(\mathcal{T})$ associates with a dataset $\mathcal{D}_i$. Each dataset $\mathcal{D}_i$ consists of a support set $\mathcal{D}^{s}_i=\{(X^s_{j},Y^s_{j})\}^{N^s}_{j=1}$ and a query set $ \mathcal{D}^{q}_i=\{(X^q_{j},Y^q_j)\}^{N^q}_{j=1} $, where $X$ and $Y$ denote the input and ground truth of a sample, respectively.
During the meta-training stage, a task-specific (a.k.a., post-update) model $f_{\theta'_i}$ is first obtained for each task $\mathcal{T}_i$ via gradient descent over its support set $ \mathcal{D}^s_i $. 
Then MAML updates its initialization (a.k.a., pre-update) $ \theta $ according to the performance of $ f_{\theta'_i} $ on the query set $ \mathcal{D}^q_i$ as in Eq.\ref{eq:maml}:
\vspace{-5pt}\begin{align}
\label{eq:maml}
\theta^{*} = \min _{\theta} {E}_{\mathcal{T}_{i} \sim p(\mathcal{T})}\left[\mathcal{L}\left(f_{\theta'_{i}}\left({X}_{i}^{q}\right), {Y}_{i}^{q}\right)\right]\\
\label{eq:maml1}
\text{s.t.}~ \theta'_{i}=\theta-\alpha \nabla_{\theta} \mathcal{L}\left(f_{\theta}\left({X}_{i}^{s}\right), {Y}_{i}^{s}\right)
\end{align}\vspace*{-18pt}

\noindent
where $ \alpha $ is the inner loop learning rate.
During the meta-testing stage, the learned initialization $ \theta^* $ is fine-tuned on the support set $ \mathcal{D}_t^s $ for task $ \mathcal{T}_t $, and the resulting model is evaluated on the query set $ \mathcal{D}_t^q $ with the post-update parameters $ \theta'_t $.

\section{Methodology} 
\begin{figure*}
 \centering 
\includegraphics[width=0.89\textwidth]{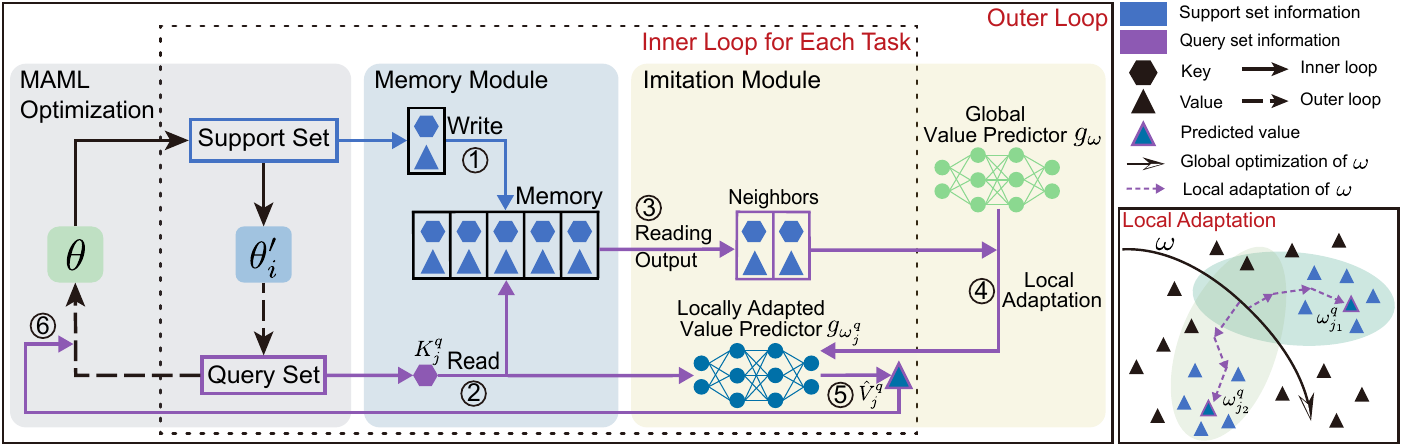}
\caption{The architecture of our model, MemIML. 
The left area details the procedure of predicting a query-set sample $X^q_j$ in each task with a task-specific memory module and an imitation module shared across tasks.
The right area illustrates the local adaption of the value predictor. The two green areas represent the neighboring areas of the global parameters $ \omega $ for two query-set samples in one task.}
\label{model:whole}
\end{figure*}

To alleviate the memorization overfitting issue in meta-learning, we propose MemIML,
which includes a memory module and an imitation module on the grounds of a base model.
The memory module is task-specific, recording the mapping behaviors between inputs and outputs of support sets for each task. 
The imitation module is shared across tasks and predicts values for each query-set sample by dynamically imitating the memory construction.
The acquired support set information leveraged by the imitation module augments the model initialization learning, enhancing the dependence of the model's task adaptation on support sets. 
Fig.~\ref{model:whole} shows our model architecture. 
\subsection{Memory Module}
\label{model:memory}
We design a memory module $M_i$ for each task $ \mathcal{T}_i$ and incorporate it in the MAML framework.
In order to fully leverage information from support sets, we construct key-value pairs from support-set samples and store them in the memory module. 
The key is the sentence representation of a sample input from support sets obtained from an introduced key network.
The corresponding value is constructed to store the information of the sample output (ground truth) as in Sec.~\ref{ssec:application}: in NLG tasks, the value is the sentence embedding of the output sentence; in NLU tasks, the value is the one hot embedding of the class label (a scalar) of the sample.
Our memory has two operations: \textit{memory writing} that constructs the memory and \textit{memory reading} that acquires information from memory. 
In the following, we elaborate on these contents in detail. 

\paragraph{Key Network} represents a sample with a vector. \label{sec:key_network} Specifically, we use a frozen pre-trained BERT model \citep{bert} as the key network.
The input of the key network is the sample input sentence $ X^s_j \in \mathcal{D}^{s}_i$ ($ X^q_j \in \mathcal{D}^q_i $), and the output is the encoded representation of the first token (i.e.\ [CLS] token) of the sentence. The acquired representation is regarded as the key $ K^s_j$ for $X^s_j$ ($K^q_j$ for $X^q_j$).
\paragraph{Memory Writing} constructs the memory using the information of samples in the support set ${\mathcal{D}^s_i}$.
\label{sec:memory_write}
For each task $ \mathcal{T}_i $, the task-specific memory $M_i$ consists of $ N^{i} $ memory slots (i.e.\ key-value pairs $\{K^s_l,V^s_l\}_{l=1}^{N_i}$). 
To build these memory slots,  we select samples from support sets and write their information into the memory. The sample selection is according to a diversity-based selection criterion \citep{diversity_metric} to ensure the diversity and representativeness of the memory content. The detailed description of this criterion is in Appendix~\ref{diversity}.

For each task-specific memory module $ M_i $, we adopt the diversity score as $ S(M_i)$ on the stored keys. 
Here, a more diverse memory gets a higher diversity score.
When the memory is not full, we directly write support-set samples without selection; otherwise, 
we compute the diversity score of the current memory and scores after every old key-value pair is replaced with a new key-value pair. Then we replace the old pair with the new one where the replacement can maximize the diversity score.
In this way, the memory we build can carry more distinguishable and representative information and efficiently utilize the storage space.

\paragraph{Memory Reading} obtains information from memory to enhance the meta-learning.     
\label{read}
The input is the sentence representation of the sample in query sets encoded by the key network, and the output is the memory slots similar to the query sample.
Specifically, given the key representation $ K^{q}_j$ of a sample $ X^q_j \in \mathcal{D}^q_i $, we retrieve the top $ N $ most similar slots from its task-specific memory $M_i$. The similarity is measured based on the Euclidean distance between $ K^q_j $ and each key $K^s_l$ in the memory slots. 
The retrieved key-value pairs $\{K^s_l, V_l^s\}_{l=1}^{N} $ act as the output of memory reading.

\subsection{Imitation Module}
\label{ssec:imitation}
In order to better leverage the retrieved memory and enhance the dependence of our model on support sets, we propose an imitation module to encourage the imitation of support sets behaviors when making predictions on query sets. 
For each sample $X^q_j$ in the query set, the inputs of the imitation module are the key $K^q_j$ and its retrieved $N$ memory slots, and the output is the predicted value $\hat{V}^q_j$ for $X^q_j$.
To achieve the imitation, we construct a value predictor that can model the behaviors of support-set samples (i.e.\ key-value matching) stored in the memory. For estimating the value of each query-set sample, we conduct \textit{local adaptation} on the value predictor
to adapt the matching.

In this way, the proposed imitation module is customized for each query-set sample, which facilitates better capture of specific task information than directly using the memory reading output, especially when tasks are versatile. The reason is that the similarity measurement of previous memory reading operations is based on the fixed BERT representations, which ignores the task-specific information.

\subsubsection{Value Predictor}
In MemIML, the proposed value predictor aims to build a mapping from keys to values of the memory module mentioned in Sec.~\ref{sec:memory_write}. 
The input of the value predictor is a key obtained from the key network, and the output is the associated value.

Specifically, we use a two-layer fully-connected network $g_\omega$ with parameters $\omega$ to build the mapping. 
The value predictor is learned over constructed key-value pairs of support sets across all tasks. 
Given the key $K^q_j$ of a query-set sample input $X^q_j$, we can then estimate its associated value as $\hat{V}^{q}_j$.

\subsubsection{Training of The Value Predictor}
To train the value predictor, we minimize the reconstruction loss $\mathcal{L}_{\omega}^{rec}(\hat{V},V)$ to make the predicted values as close as possible to values constructed from the ground truths of support-set samples, where $\mathcal{L}^{rec}_{\omega}$ is the cross-entropy loss if the value $V$ is a label and is the mean square loss if $V$ is a vector.

The training procedure includes the global optimization shared across tasks and the local adaptation for each specific task. 
Specifically, we first train the value predictor with samples from support sets of all tasks. After feeding the memory reading output of a query-set sample to this network, we perform local adaptation and employ the adapted  network to estimate the value for the query sample.

\paragraph{Global Optimization.}
\label{sec:global_optimization}
To obtain the task-independent global parameters $\omega$, we train the value predictor over constructed keys (i.e., \ as inputs) and values (i.e., \ as outputs) from support-set samples of all tasks. 
The global optimization keeps updating in the whole meta-training phase.

\paragraph{Local Adaptation.}
\label{LA}
To make the value predictor adaptive to each query-set sample $X^q_j$, inspired by \citep{mbpa}, we propose local adaptation that fine-tunes the global value predictor $g_{\omega}$ to get an adapted one with parameters $ \omega^q_j$. The local adaptation only works when predicting $X^q_j$.

Based on the initial parameters $\omega$ from the global optimization, we perform several gradient descent steps to minimize the loss $\mathcal{L}^{loc}$, which is:
\vspace{-5pt}\begin{align}
\mathcal{L}^{loc}= \gamma\|\tilde{{\omega}}-{\omega}\|_{2}^{2} + \frac{1}{N}\sum_{l=1}^{N} \mathcal{L}^{rec}_{\tilde{\omega}}(\hat{V}^s_l,V^s_l)
\label{loss:la}
\end{align}\vspace{-12pt}

\noindent
Here, $\hat{V}^s_l=g_{\tilde{\omega}}(K^s_l)$, $\{K_l^s,V_l^s\}_{l=1}^{N}$ is the memory reading output of the query-set sample, and the factor $ \gamma $ restricts the distance between $\omega^q_j$ and $\omega$. 
Minimizing the second term encourages $g_{\omega^q_j}$ to better estimate the retrieved memory values $\{V^s_l\}_{l=1}^{N}$. 
Then we can acquire the locally adapted value prediction network $g_{\omega^q_j}$ with parameters $\omega^q_j=\underset{\tilde{{\omega}}}{\arg \min} \mathcal{L}^{loc}(\tilde{\omega})$. 
Given a query-sample key $K^q_j$, we can thus predict its associated value as
\vspace{-7pt}\begin{align}
\hat{V}^q_j=g_{\omega^q_j}(K^q_j),
\end{align}\vspace{-19pt}

\noindent
where the adapted parameters $\omega^q_j$ are discarded thereafter, and the model does not back-propagate through $\hat{V}^q_j$.

In this sense, besides the task-specific parameter $\theta'_i$ provided by MAML, there will also be $\omega^q_j$ learned from support sets specific to each query-set sample. This guarantees that the model relies more on support sets for task adaptation.
Fig.~\ref{model:whole} (right part) illustrates the mechanism of local adaptation. 

\subsection{MemIML on NLP Applications}
\label{ssec:application}
In this part, we will elaborate on two few-shot applications in NLP (i.e., \ text generation and text classification) to solve the memorization overfitting problem of MAML. The model structures of these applications are basically the same, except for the following three points:
the base model, the way to get the value $V^s_l$ stored in the memory module, and the way to leverage the output $\hat{V}^q_j$ of Sec.~\ref{ssec:imitation}.
\label{application}
\paragraph{Personalized Dialogue Generation.} 
The base model is the transformer \citep{transformer} consisting of an encoder and a decoder. In this task, each sample consists of an input utterance and a ground truth utterance, so the value $V^s_l$ stored in the memory is obtained from the ground truth utterance $ Y^s_l$ of a support-set sample, which is embedded by the key network followed by an LSTM~\citep{lstm}. This LSTM is optimized with the base model.
The $\hat{V}^q_j$, concatenated with the encoder outputs, serves as a new input for the decoder. 
Hence, we acquire the prediction of a query-set sample via $\hat{Y}^q_j=\text{Decoder}([\hat{V}^q_j;\text{Encoder}(X^q_j)])$.

\paragraph{Multi-domain Sentiment Classification.}
The base model is a BERT \citep{bert} followed by a fully-connected network.
Each sample consists of an input sentence and a sentiment label (ground truth), so the memory value $V^s_l$ is the sentiment label. 
To leverage $\hat{V}^q_j$, we interpolate it with the original output of the base model $ \tilde{Y}^q_j $ as 
\vspace*{-5pt}\begin{align} 
  \hat{Y}_j^{q}=\beta \tilde{Y}_j^q + (1-\beta) \hat{V}_j^q
  \label{eq:nlu_prediction}
\end{align} 
\vspace*{-20pt}

\noindent where $ \beta $ balances $ \tilde{Y}^q_j $ and $ \hat{V}^q_j$. Notice that the interpolation not only works on the prediction output but also guides the training via gradient descent based on the interpolated output. We verify the effectiveness of the interpolation in Appendix~\ref{Exp:nomm}.

\begin{algorithm}[h]
  \algsetup{linenosize=\tiny} 
  \small
  \caption{Memory Imitation Meta-training}
  \label{al:train}
  \begin{algorithmic}[1]
    \REQUIRE $p(\mathcal{T}) $: task distribution,
     $\alpha_{1-4}$: step sizes
    \STATE Initialize $\theta$ from pretrained model; initialize $ \omega $ randomly; initialize memory for $T$ tasks as $\{{M}_i \}_{i=1}^{T} =\{ \upphi \}_{j=1}^{T}$
    \WHILE {not converge}
    \STATE Sample batch of tasks $\{\mathcal{T}_{i}\}_{i=1}^{n} $, where $ \mathcal{T}_i  \sim p(\mathcal{T}) $
    \FORALL {task $\mathcal{T}_{i} $}
    \STATE Sample support set $\mathcal{D} _{i}^{s}$ and query set $ \mathcal{D}_i^q$ from $\mathcal{T}_{i}$
    \STATE Obtain the keys $\{K_l^s\}_{l=1}^{N^s}$ and the values $ \{V_l^s\}_{l=1}^{N^s} $ for the support set $ \mathcal{\mathcal{D}}_i^s $ as in Sec.~\ref{sec:key_network}
    \STATE $ M_i \leftarrow \{<K_l^s,V_l^s>\}_{l=1}^{N^s} $ \# Write memory 
    \STATE $ \omega \leftarrow \omega - \alpha_1 \nabla_\omega \mathcal{L}^{rec}$  \# Global optimization
    \STATE$ \theta'_i \leftarrow \theta - \alpha_2 \nabla_\theta  \mathcal{L}^{base}$ \# Learn $\theta'_i$ in Eq.~\ref{eq:maml1}
    \FOR {$(X^q_j,Y^q_j) \; \text{in} \; \mathcal{D}_i^q$} 
    \STATE Obtain the keys $ K^q_j $ for each sample $X^q_j$
    \STATE Retrieve $ N $ nearest neighbors of $ K^q_j $ from $ {M}_i $. 
    \STATE $ \omega^q_j \leftarrow \omega - \alpha_{3} \nabla_{\omega} \mathcal{L}^{loc} $ \# Local adaptation
    \STATE $ \hat{V}^q_j=g_{\omega^q_j}(K^q_j) $ \# Predict memory output
    \STATE Predict $ \hat{Y}^q_j $ as in  Sec.~\ref{application}
    \ENDFOR
    \ENDFOR
    \STATE Update $\theta \leftarrow \theta -\alpha_4 \nabla_{\theta} \sum_{\mathcal{T}_i \sim p(\mathcal{T}) } \mathcal{L}^{{base}}_{\mathcal{T}_i,\theta'_i} (\hat{Y}^{q}, {Y}^{q})  $ 
    \ENDWHILE
  \end{algorithmic}
  \end{algorithm}
  
 \begin{table*}[htbp]
  \setlength\tabcolsep{2.2pt} 
  \renewcommand\arraystretch{1.1}
  \centering
  \scalebox{0.73}{
  \begin{tabular}{c|c|cccc|c|c|cccc|c|c|c}
  \Xhline{3\arrayrulewidth}
  \multirow{3}{*}{Methods} & \multicolumn{12}{c}{\textbf{\textit{Automatic Metrics}} }  & \multicolumn{2}{|c}{\multirow{2}{*}{\textbf{\textit{Human Evaluation}}} }    \\ 
  \cline{2-13}
  & \multicolumn{7}{c|}{\textit{Quality}} & \multicolumn{4}{c|}{\textit{Diversity}} & \textit{Consistency} &  \multicolumn{2}{c}{}\\
  \cline{2-15} & {PPL}  & {BLEU1} & {BLEU2} &{BLEU3}& {BLEU4}& {ROUGE}& {CIDEr} &{Dist1}&  {Dist2}& {Dist3} & {Dist4} &{C-score} & {Quality} & {Consistency}\\
  \hline
  Base Model    & 38.14   & 15.53 & 6.810  & 3.430  & 1.948 & 0.163 & 0.136  & 0.006  & 0.023  & 0.048  & 0.080   & -0.024 & 0.689  & 0.395  \\
  Fine-tune     & \textbf{34.14}  & 16.10 & 7.222  & 3.678  & 2.100 & 0.166  & 0.147  & 0.007 & 0.028  & 0.063  & 0.111    & 0.012 &0.886  &0.641  \\
  MAML & 43.24   & 15.56 & 7.456  & 3.858  & 2.229  & 0.172 & 0.152  & 0.013 & 0.046  & 0.099  & 0.169    & 0.156 &0.807  & 0.651 \\
  MR-MAML      & 52.52   & 13.35 & 5.571  & 2.783  & 1.601 & 0.142 & 0.110  & 0.004 & 0.011  & 0.021   & 0.034   & 0.132 & 0.512 & 0.562   \\
  \hline
  MemIML & { 41.61} & {\textbf{16.23*}} & {\textbf{7.941*}} & {\textbf{4.295*}} & {\textbf{2.557*}}  & \textbf{0.183*} & \textbf{0.173*} & {\textbf{0.014*}} & {\textbf{0.053*}} & {\textbf{0.114*}} & {\textbf{0.195*}} & { \textbf{0.241*}} & \textbf{0.932}  &\textbf{0.807} \\
  \Xhline{3\arrayrulewidth}
  \end{tabular}}
  \caption{Overall performance over Persona-Chat dataset. The results with * indicate that the improvements of our model overall baselines are statistically significant with $p<0.05$ under t-test.}
  \label{table:persona}
  \end{table*}  
\subsection{Theoretical Analysis}  
We theoretically investigate how our method helps to alleviate the memorization overfitting problem. 
Following \citet{mm_overfit}, we use mutual information $\mathcal{I}(\hat{Y}_i^q;\mathcal{D}_i^s|\theta,X_i^q)$ to measure the level of the memorization overfitting. When the learned model ignores support sets to predict query sets, $\mathcal{I}(\hat{Y}_i^q;\mathcal{D}_i^s)|\theta,X_i^q)=0$ occurs, which indicates the complete memorization overfitting in meta-learning \citep{mm_overfit}. Hence, lower mutual information means more serious memorization overfitting issues.

We propose a criterion similar to \cite{task_aug} to measure the validity of our method for tackling this problem.
For a task $\mathcal{T}_i=\{D^s_i,D^q_i\}$, the criterion aims to mitigate the memorization overfitting by enhancing the model's dependence on the support set $ \mathcal{D}^s_i $, i.e.\ increasing the mutual information between support set and $ \hat{Y}^q_i $ as follows:
\vspace{-5pt}\begin{align}
    \mathcal{I}(\hat{Y}^{q}_i; \![\mathcal{D}^s_i, \mathcal{M}_i]\!\mid\! \theta,X^{q}_i)\!>\!\mathcal{I}(\hat{Y}^{q}_i ; \mathcal{D}_i^{s} \!\mid\! \theta, X^{q}_i),
    \label{eq:mul_info}
\end{align}\vspace{-18pt}

\noindent
where $\mathcal{M}_i $ means additional memory information we provide, which contains support sets information to augment the inference of the sample $ X^q_i $ in $\mathcal{D}^{q}_i$.
We demonstrate our method MemIML meets the above criterion (See details in Appendix~\ref{proof}.).

\subsection{The Procedure of Training and Testing}
In the meta-training phase (shown in Alg.~\ref{al:train}), MemIML first constructs an empty memory for each task and then follows the bi-level optimization process of MAML. 
In the inner loop, MemIML adapts the base model initialization $\theta$ to task-specific parameters via training on the support set. At the same time, from each support-set sample, MemIML obtains a key-value pair and determines whether to write it into the memory or not. Then, MemIML conducts the global optimization of the value predictor over these key-value pairs. In the outer loop, each sample of the query set reads the memory to retrieve the most similar memory slots. 
Local adaptation fine-tunes the value predictor on those retrieved slots. Next, the adapted value predictor estimates the value of each query sample and uses it to augment the learning of the model initialization. 
The total loss function in the inner loop is $\mathcal{L}^{total}=\mathcal{L}^{{base}} + \mathcal{L}^{{rec}}$,
where $ \mathcal{L}^{{base}}=\mathcal{L}(f(X^{s}),Y ^{s})$ is the cross-entropy loss.

The procedure of meta-training and meta-testing are almost the same except that meta-testing does not optimize the learned model initialization $\theta$ and the initial parameter $\omega$ of the value predictor. For each task $\mathcal{T}_t$ in the meta-testing phase, MemIML also adapts $\theta$ to task-specific parameters $\theta'_i$ in the inner-loop and constructs the task-specific memory. In the outer-loop, MemIML retrieves key-value pairs from the memory to conduct local adaptation based on the initial parameter $\omega$. The estimated value $\hat{V}^q_t$ from local adaptation helps the base model to infer the final output $\hat{Y}^q_t$.

\section{Experiments and Analysis}
Experiments on personalized dialogue generation and multi-domain sentiment classification verify our model on text generation and classification, respectively, where we use Persona-Chat and ARSC datasets. 

\subsection{Personalized Dialogue Generation}
\paragraph{Dataset.} Following \citep{zhang2018personalizing}, we use Persona-chat \citep{paml} by regarding building a dialog model for each person as a task. 
The dataset consists of a training/validation/testing set with 1137/99/100 persons (tasks) separately.
In the Persona-Chat dataset, each persona description has 8.3 unique dialogues on average, and each task consists of three samples.

\paragraph{Baselines.}
We compare our methods with the following baselines:
\textbf{Base Model}: We pretrain a conventional transformer-based dialog generation model over all the training tasks ignoring the speakers' personality.  
\textbf{Fine-tune}: We fine-tune the pre-trained base model on the support sets of each meta-testing task.
\textbf{MAML}: We apply MAML~\citep{paml} to the base model. \textbf{MR-MAML}: \citet{mm_overfit} tackle the memorization overfitting of MAML via regularization.

\paragraph{Metrics.}
Automatic evaluation has three aspects,
\begin{itemize}[leftmargin=*]
\setlength{\itemsep}{0.5pt}
\setlength{\parsep}{0pt}
\setlength{\parskip}{0pt}
  \item \textit{Quality}: \textbf{BLEU-n} \citep{bleu}, \textbf{CIDEr} \citep{CIDEr}, and \textbf{ROUGE} \citep{rouge} measures the n-gram matching between the generated response and ground truth. \textbf{PPL} (perplexity) measures the sentence fluency.
  \item \textit{Diversity.} \textbf{Dist-n} \citep{li2016diversity} evaluates the response diversity by counting unique n-grams. 
  \item \textit{Consistency}: \textbf{C score} \citep{paml} measures the consistency between the generated responses and persona descriptions through a pretrained natural language inference model. 
\end{itemize}
Human evaluation consists of \textbf{{Quality}} and \textbf{{Consistency}}. (See details in Appendix~\ref{human}).

\begin{table}[!t]
  \centering
  \resizebox{7.3cm}{!}{
\begin{tabular}{c|c|c}
  \Xhline{3\arrayrulewidth}
Type &  Methods & Accuracy \\
  \hline 
Non meta-learning & Fine-tune  & 80.73 \\
\hline
 & Matching Net  & 81.22 \\
Metric-based & Prototypical Net  & 80.13 \\
 & Proto ++   &  82.41  \\
meta-learning & Relation Net  & 81.32 \\
 & Induction Net  & 79.31 \\
 \hline
 & MAML  &  82.17  \\
Optimization-based & MR-MAML  & 78.14  \\
 & Meta-Aug &  83.57  \\
meta-learning & MetaMix &  83.63  \\
  \cline{2-3}
 & MemIML (Ours) &  \textbf{85.69*} \\
\Xhline{3\arrayrulewidth}
\end{tabular}}
  \caption{
  The results of mean accuracy
  over the ARSC. * indicates that our
  improvement overall baselines is
  statistically significant with $p<0.01$
  under t-test.}
  \label{table:arsc_res} 
\end{table}

\begin{figure*}[!t]
 \centering
 \begin{subfigure}[b]{.29\linewidth}
  \centering
  \includegraphics[width=\textwidth]{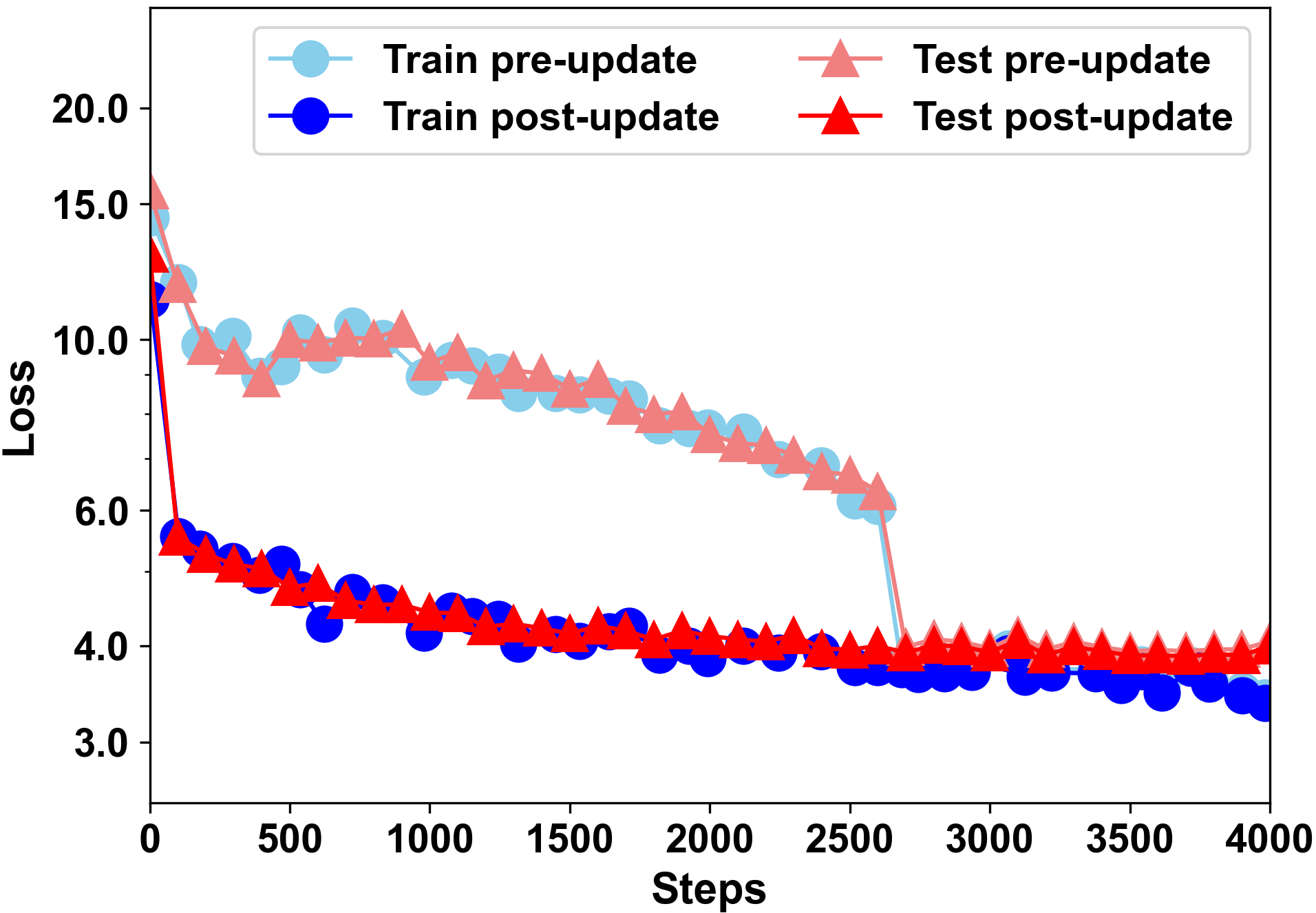}
  \caption{MAML}
 \end{subfigure}
 \begin{subfigure}[b]{.29\linewidth}
  \centering
  \includegraphics[width=\textwidth]{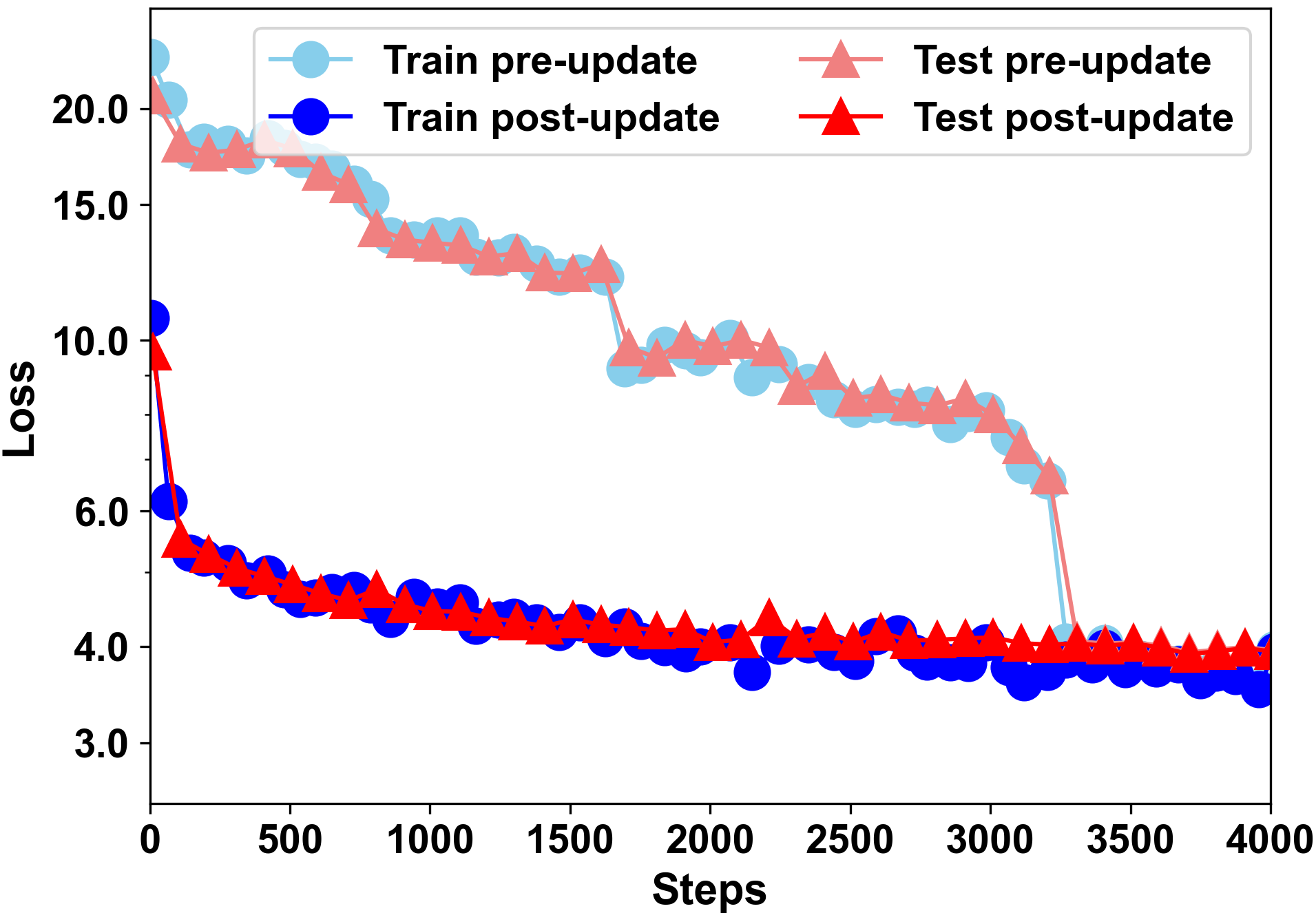}
  \caption{MR-MAML}
 \end{subfigure}
 \begin{subfigure}[b]{.29\linewidth}
  \centering
  \includegraphics[width=\textwidth]{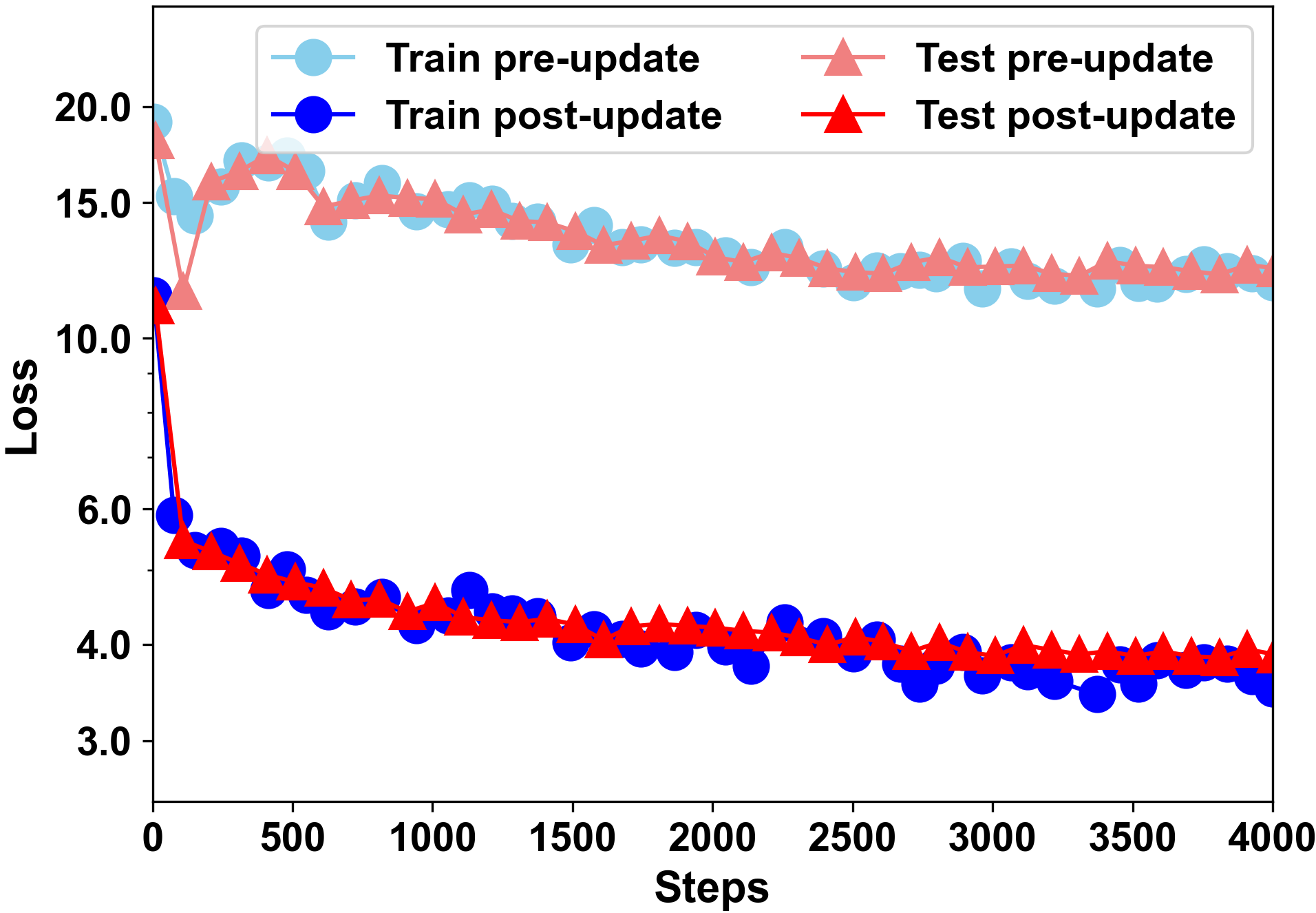}
  \caption{MemIML (Ours)}
 \end{subfigure}
 \caption{Memorization overfitting analysis on Persona-Chat. Small loss gaps between pre-update $\theta$ and post-update $\theta'_i$ (in MAML and MR-MAML) indicate the serious memorization overfitting issue (i.e., \ the gap between sky-blue and blue curves in meta-training and the gap between pink and red curves in meta-testing). The large gap in MemIML demonstrates the effectiveness of our method.}
 \label{loss_curve}
\end{figure*}

\paragraph{Overall Performance.}
As shown in Table~\ref{table:persona}. \textbf{Fine-tune} outperforms \textbf{Base Model} in all metrics, which verifies that the task-specific data is helpful to its performance on specific tasks. Compared to \textbf{Fine-tune}, \textbf{MAML} behaves better on diversity and consistency but behaves worse on quality. 
Pretraining the base model achieves the best perplexity (lowest PPL) as shown by \textbf{Base Model} and \textbf{Fine-tune}. We analyze that it's because pretraining leads to a considerable degree of fluency in their generated utterances and is careless about each task's specific information, resulting in low consistency with tasks. Our model, \textbf{MemIML}, performs the best in most aspects, including quality, diversity, and task consistency. In particular, MemIML significantly improves MR-MAML in alleviating the memorization overfitting issue, suggesting that memory imitation is more effective than only regularizing model initialization. 

\subsection{Multi-domain Sentiment Classification}
\paragraph{Dataset.} 
Amazon Review sentiment classification dataset (ARSC) \citep{arsc} contains 69 tasks in total. 
Following \citep{induction}, we build a 2-way 5-shot meta-learning with 57 tasks for meta-training and 12 tasks for meta-testing.
We conduct experiments on the ARSC \citep{arsc}. It contains English reviews of 23 types of Amazon products, where each product consists of three different binary classification tasks.
Following \citet{induction}, we select 12 tasks from 4 domains (\textit{Books, DVD, Electronics, Kitchen}) for meta-testing tasks, and the support sets of these tasks are fixed \citep{arsc}.
\paragraph{Baselines.} 
We compare our methods with the following baselines:
\textbf{Fine-tune}: We fine-tune a pre-trained BERT on the support set of meta-testing tasks (non-meta-learning method) as in Appendix~\ref{Exp:nlu}.
We choose five metric-based meta-learning baselines:
\textbf{Matching Net} \citep{vinyals2016matching},
\textbf{Prototypical Net} \citep{snell2017prototypical}, \textbf{Proto ++}, \citep{ren2018meta},
\textbf{Relation Net} \citep{sung2018learning}, and \textbf{Induction Net} \citep{induction}.
We apply an optimization-based baseline (\textbf{MAML})~\citep{maml} to the base model,
and implement some approaches tackling the memorization overfitting problem based on MAML: \textbf{MR-MAML} \citep{mm_overfit},
\textbf{MetaMix}, \citep{task_aug} and \textbf{Meta-Aug} \citep{meta_aug}.

\paragraph{Overall Performance.}
Table~\ref{table:arsc_res} shows the performance measured by the mean accuracy of meta-testing tasks.
Our model, \textbf{MemIML} outperforms all competing approaches including non-meta-learning, metric-based meta-learning, and optimization-based meta-learning methods. Particularly, our model surpasses the current solutions to the memorization overfitting problem (\textbf{MR-MAML, Meta-Aug, MetaMix}), indicating that our method is more effective compared to regularization and textual augmentation.

\begin{table*}[htbp]
    \centering
    \parbox{0.6\textwidth}{
    \scalebox{0.73}{
  \begin{tabular}{@{}l@{\hspace{1pt}}|@{\hspace{1pt}}c@{\hspace{1pt}}|@{\hspace{1pt}}c@{\hspace{1pt}}|@{\hspace{1pt}}c @{\hspace{1pt}}|@{\hspace{1pt}}c@{\hspace{1pt}}|@{\hspace{1pt}}c @{\hspace{1pt}}|@{\hspace{1pt}}c@{\hspace{1pt}}|@{\hspace{1pt}}c@{\hspace{1pt}}|@{\hspace{1pt}}c@{\hspace{1pt}}|@{\hspace{0pt}}c@{}}
  \Xhline{3\arrayrulewidth}
        & \multicolumn{8}{@{}c@{}|}{\textbf{Persona-Chat}} & \textbf{ARSC} \\ \cline{2-10} 
        & PPL   & C-score & BLEU3 & BLEU4 & Dist1  & Dist2  & ROUGE & CIDEr & Acc     \\ \hline
  MemIML  & \textbf{41.62} & \textbf{0.240}       & \textbf{4.295} & \textbf{2.557} & \textbf{0.014} & \textbf{0.053} & \textbf{0.183} & \textbf{0.173} & \textbf{85.69}     \\
  \hline
  - Similarity-Search & 45.17 & 0.153       & 3.817 & 2.219 & 0.011 & 0.044 & 0.168 & 0.158 & 84.14     \\
  - Value predictor & 42.93 & 0.183       & 4.199 & 2.313 & 0.010 & 0.039 & 0.182 & 0.167 & 84.67   \\
  - Local Adaptation  & 48.08 & -0.117      & 3.452 & 1.948 & 0.007 & 0.023 & 0.171 & 0.129 & 84.19        \\
\Xhline{4\arrayrulewidth}
  \end{tabular}}
  \caption{Ablation Studies. - means deleting MemIML's components.}
  \label{Ab:all}
  }
    \parbox{0.39\textwidth}{
  \centering
  \scalebox{0.73}{
  \begin{tabular}{@{}c@{\hspace{3pt}}|@{\hspace{3pt}}c@{\hspace{3pt}}|@{\hspace{3pt}}c@{\hspace{3pt}}|@{\hspace{3pt}}c@{}}
  \Xhline{3\arrayrulewidth}
  \multicolumn{4}{c}{\textbf{Memory Analysis on ARSC}}\\\hline
    {Store ratio} & {Acc} & {\# Neighbors} &  Acc \\ \hline
    {100\%}   & {84.91}        & {5}    & 84.04 \\
    {80\%} & {85.69}        & {10}   & 84.47 \\
    {50\%} & {84.84}        & {20}   & 85.69 \\
    {20\%} & {84.35}        & {50}   & 85.04\\
  \Xhline{3\arrayrulewidth}
  \end{tabular}}
  \caption{Memory analysis on ARSC.}
  \label{table:Aarsc}
}
\end{table*}
\subsection{Memorization Overfitting Analysis}
In Figure~\ref{loss_curve}, the gaps of the losses on query sets between pre-update $\theta$ (before training on support sets) and post-update $\theta'_i$ (after training on support sets) indicate the memorization overfitting problem. The gap between sky-blue and blue curves measures the memorization overfitting of meta-training (the gap between pink and red curves measures meta-testing). Small loss gaps indicate a severe memorization overfitting where support sets are almost useless for task adaptation.
Those loss gaps between $\theta$ and $\theta'_i$ collapse in MAML and MR-MAML after about 3000 steps. This indicates that the post-update $\theta'_i$ barely benefits from the support set, and thus the memorization overfitting issue is severe. 
In Figure~\ref{loss_curve} (c), MemIML has large gaps between $\theta$ and $\theta'_i$, implying that $\theta'_i$ better leverages support sets when adapting to new tasks and thus alleviates the memorization overfitting issue.

\subsection{Ablation Studies}
In Table~\ref{Ab:all}, we conduct ablation studies to verify the effectiveness of each component.   
Removing \textit{Similarity-Search} means the memory reading operation randomly outputs memory slots instead of searching for similar memory slots. This variant underperforms MemIML, indicating that similar samples stored in the memory provide more useful information to improve the model performance.
Removing the \textit{value predictor} means directly using the memory output without a learnable network.
Its results are not too bad, indicating that the memory module helps to mitigate the memorization overfitting problem. However, this usage simply aggregates the support set information into the query set, which is not as precise as learning the information required by the query set itself. Therefore, 
it is still inferior to our model. 
Removing \textit{Local adaptation} means we only use the global value predictor to estimate the memory output. It is crucial to the value predictor since removing it from the value predictor results in an even worse performance than removing the \textit{value predictor}. 
Besides, the significant drop in task consistency (C-score) shows that local adaptation contributes a lot to making the model adaptive to specific tasks, as it learns to adapt to each query-set sample. 

\subsection{Analysis of Memory Operations}
\paragraph{Memory Size.}
In Table~\ref{table:Aarsc} and \ref{table:Aperson}, we investigate the variants of our task-specific memory module of different sizes. We control the memory size through $|M|=\text{store ratio}\times |D^s|$.
The results demonstrate that our model is able to maintain high performance even with only a 20\% memory size by storing diverse and representative samples of support sets. 
Besides, as the ratio of stored samples increases, the model's performance is improved since  it provides more information for the inference of query samples and the optimization of the model initialization. Storing all the encountered samples (i.e., \ with store ratio 100\%) in the memory instead introduces some noise that damages the model performance.

\paragraph{Number of Neighbors.}We also investigate the effects of different numbers of neighbors for the model performance in Table~\ref{table:Aarsc} and Table~\ref{table:Aperson}.
In both datasets, the model performs better with a larger number of neighbors. However, when the number of neighbors is too large, the model retrieves some dissimilar slots from the memory module.
These dissimilar slots bring much noise, which makes the predictions of query samples inaccurate.

\begin{table}[!t]
  \centering
  \setlength\tabcolsep{2pt} 
  \renewcommand\arraystretch{1.1}
  \resizebox{7.7cm}{!}{
  \begin{tabular}{c|c|c|c|cc|cc|c|c}
  \Xhline{3\arrayrulewidth}
     \multicolumn{2}{c|}{ }&    PPL   & C-score & BLEU3 & BLEU4 & Dist1 & Dist2 & ROUGE & CIDEr \\ \hline
& 1   & 43.54 & 0.197  & 4.224  & 2.447  & 0.014 & 0.055 & 0.179 & 0.174 \\
Store     & 0.8 & 43.21 & 0.198  & 4.414  & 2.622  & 0.014 & 0.054 & 0.182 & 0.183 \\
  ratio & 0.5 & 41.86 & 0.223  & 4.069  & 2.317  & 0.013 & 0.052 & 0.179 & 0.162 \\
          & 0.2 & 41.97 & 0.204  & 4.021  & 2.271  & 0.012 & 0.052 & 0.181 & 0.168 \\
  \hline
  \hline
   & 5   & 41.98 & 0.192  & 3.855  & 2.203  & 0.013 & 0.053 & 0.177 & 0.162 \\
   Neighbor      & 10  & 41.62 & 0.239  & 4.295  & 2.557  & 0.014 & 0.053 & 0.183 & 0.173 \\
    number       & 20  & 42.12 & 0.155  & 4.099  & 2.336  & 0.012 & 0.046 & 0.179 & 0.165 \\
          & 50  & 42.76 & 0.145  & 3.614  & 2.072  & 0.011 & 0.041 & 0.169 & 0.144 \\
\Xhline{3\arrayrulewidth}
  \end{tabular}}
  \caption{Analysis of Persona-Chat dataset.}
  \label{table:Aperson}
  \end{table}
  \raggedbottom
\subsection{Case Study}
We present two generated cases in personalized dialog in Table.~\ref{case}.
\textbf{Base Model}, \textbf{Fine-tune}, and \textbf{MAML} generate general responses with little useful information or responses that are not consistent with the personality of personas. \textbf{MR-MAML} generates irrelevant responses to the dialogue context. 
Our model not only responds coherently to the dialog history but also caters to the persona descriptions of each user. 

\begin{table}[ht]
\centering
\scalebox{0.71}{
  \begin{tabular}{p{0.13\textwidth}|p{0.5\textwidth}}
\Xhline{3\arrayrulewidth}
  Persona A     & \begin{tabular}[c]{@{}l@{}} I am a professional singer.\\ I graduated \textcolor{cyan}{from USC}.\\ 
  I love watching \textcolor{blue}{horror films}.\\ 
  I was born in South Carolina.\end{tabular}   \\ \hline
  Dialogues     & 
  \begin{tabular}[c]{@{}p{1.3mm}p{0.45\textwidth}@{}}A: & Where are you from?  I grew up in South Carolina.\\ 
  B: & I was born in Arkansas, but I now live in Chicago. \\ 
  A: & Did you go to college there? \\ 
  B: & Yes, I am in school now while working part-time as a barista. 
  \end{tabular} \\ \hline\hline
  Human Response      & I graduated recently \textcolor{cyan}{from USC}.My parents pushed me into it.       \\ \hline
  Base Model & What do you do for a living?       \\ \hline
  Fine-tune      & Are you in trouble?   \\ \hline
  MAML  & Do you work in the fitness industry?        \\ \hline
  MR-MAML       & That sounds illegal. Are they not protected? \\ \hline
  MemIML  & That would be cool! I love \textcolor{blue}{horror films}. \\    
  \Xhline{3\arrayrulewidth}\hline
  \Xhline{2\arrayrulewidth}
  Persona A     & \begin{tabular}[c]{@{}l@{}} I have a large \textcolor{teal}{garden}.\\ My hobby is walking.\\ 
  I love to \textcolor{blue}{travel}.\\ 
  I \textcolor{cyan}{teach biology} at a college.\end{tabular}   \\ \hline
  Dialogues     & \begin{tabular}[c]{@{}p{1.5mm}p{0.45\textwidth}@{}}B: & Hi, Merna here. Single, 32 and scared of dogs. You?\\ 
  A:& I am dave, 40. \\ 
  B: &Oh, okay. I am a good cook! \\ 
  \end{tabular} \\ \hline\hline
  Human Response      & Cool, I grow my own vegetables in my garden \textcolor{teal}{garden}. \\ \hline
  Base Model & What do you do for a living?       \\ \hline
  Fine-tune      & What do you teach? \\ \hline
  MAML  & What do you do?     \\ \hline
  MR-MAML       & I am doing great. How are you tonight?        \\ \hline
  MemIML  & Oh, i am a \textcolor{cyan}{science teacher} and \textcolor{blue}{love travel}. \\ 
  \Xhline{3\arrayrulewidth}
  \end{tabular}
  }
  \caption{Two generated examples in the Persona-Chat dataset. Colored texts indicate that the  responses match the personality descriptions. }
  \label{case}
\end{table}

\section{Conclusion}
In this paper, we tackle the memorization overfitting problem of meta-learning for text classification and generation applications. We propose MemIML to enhance the dependence of the model on the support sets for task adaptation.
MemIML introduces a memory module storing the information of support sets, 
and propose an imitation module to better leverage the support set information by imitating the behaviors of the memory. 
Both empirical and theoretical results demonstrate that our method MemIML effectively alleviates the memorization overfitting problem.

\section{Ethical Considerations}
The persona-based dialogue generation task aims to build a dialogue model which generates meaningful, fluent, and consistent responses. It will facilitate human-computer interactions in practice. However, the training of the model for personalized dialogues may lead to the leakage of personal privacy information. 
In this work, the data source we use is from a published dataset and does not involve privacy issues for the data collection. Our proposed method does not include inference or judgments about individuals and does not generate any discriminatory, insulting responses. 
Our work validates the proposed method and baseline models on human evaluation which involves manual labor. We hire five annotators to score 750 generated sentences in total (250 sentences for each model we evaluate). The hourly pay is set to 15 US\$ per person, which is higher than the local statutory minimum wage.

\section*{Acknowledgements}
Research on this paper was supported by Hong Kong Research Grants Council (Grant No. 16204920) and National Natural Science Foundation of China (Grant No. 62106275).

\bibliography{anthology,custom}
\bibliographystyle{acl_natbib}
\newpage
\appendix
\section{Validity of Memory Imitation Strategy}
\label{proof}
\begin{proof}[Proof of inequality in Eqn.~\ref{eq:mul_info}]
  We check the validity of memory imitation by examining whether the criterion in Section 4.4 is met. We check the increase of mutual information between predictions of query sets with the provided support-set information after augmented with the memory information $ \mathcal{M} $. 
  \begin{align}
  &\mathcal{I}(\hat{Y}^q;[D^s,\mathcal{M}]|\theta,X^q) - \mathcal{I}(\hat{Y}^q;D^s|\theta,X^q)
  \nonumber \\
  ={}&H(\hat{Y}^q|\theta,X^q)-H(\hat{Y}^q|D^s,\mathcal{M},\theta,X^q)
  \nonumber \\
	 &-H(\hat{Y}^q|\theta,X^q) + H(\hat{Y}^q|D^s,\theta,X^q)\nonumber  \\
	 ={}& -H(\hat{Y}^q|X^q,X^s,Y^s,\mathcal{M},\theta)\nonumber\\
	 &+H(\hat{Y}^q|X^q,X^s,Y^s,\theta).
  \label{eq:KL_div}
  \end{align} 
  For short, we use notation $\boldsymbol{Z}=
  (X^q,X^s,Y^s,\theta )$ to denote a
  set of variables. Then we can rewrite (\ref{eq:KL_div}) as
  \begin{align*}
	  &-H(\hat{Y}^q|\boldsymbol{Z},\mathcal{M})+H(\hat{Y}^q|\boldsymbol{Z})\\
      ={}&E_{\hat{Y}^q,\boldsymbol{Z}, \mathcal{M}}\left[ \log p(\hat{Y}^q|\boldsymbol{Z},\mathcal{M}) \right] \\
      &-
	  E_{\hat{Y}^q,\boldsymbol{Z}}\left[ \log p(\hat{Y}^q|\boldsymbol{Z}) \right].
  \end{align*}
  Note that trivially, we have $E_{\mathcal{M}}\left[ 1 \right] =1$, so we get
  \[
  	E_{\hat{Y}^q, \boldsymbol{Z}}\left[ p(\hat{Y}^q|\boldsymbol{Z}) \right] =E_{\hat{Y}^q, \boldsymbol{Z}, \mathcal{M}}\left[ p(\hat{Y}^q|\boldsymbol{Z}) \right]
  \]
  since $p(\hat{Y}^q,\boldsymbol{Z})$ does not rely on the variable $\mathcal{M}$.
  Hence, we can just write $E_{\hat{Y}^q, \boldsymbol{Z}, \mathcal{M}}$ as $E$ for short. Then the equation (\ref{eq:KL_div}) will become to 
  \begin{align*}
  & E[\log p(\hat{Y}^q| \boldsymbol{Z}, \mathcal{M})]-E[\log p(\hat{Y}^q|\boldsymbol{Z})]\\
	  ={} & E[\log
	  \frac{p(\hat{Y}^q|\mathcal{M},
	  \boldsymbol{Z})}{p(\hat{Y}^q|
  \boldsymbol{Z})}] \\
  	={} & \!\!\!\!\!\!
	\sum_{\hat{Y}^q,\mathcal{M},
	\boldsymbol{Z}}^{}\!\!\!\!
	p(\boldsymbol{Z})p(\hat{Y}^q,\mathcal{M}|
	\boldsymbol{Z})\log
	\frac{p(\hat{Y}^q,\mathcal{M}|
	\boldsymbol{Z})}{p(\hat{Y}^q|
\boldsymbol{Z})p(\mathcal{M}|
\boldsymbol{Z})}\\
		={}&E_{\boldsymbol{Z}}[KL(p(\mathcal{M},
		\hat{Y}^q|\boldsymbol{Z})|\!|p(\hat{Y}^q
		|\boldsymbol{Z})p(\mathcal{M}|
		\boldsymbol{Z}))]\\
		>{}& 0
  \end{align*}
where the last inequality holds due to $ \hat{Y}^q $ is dependent on $ \mathcal{M}$.
\end{proof}

We also investigate that memory imitation improves the learning of model initialization via another criterion $ \mathcal{I}(\theta;[D^q,\mathcal{M}]|D^q) >0$ following \citet{task_aug}.
This criterion guarantees that the additional memory knowledge contributes to updating the initialization in the outer loop.
Since all the meta-training tasks satisfy this criterion, the generalization ability of the model initialization improves. 
\begin{proof}
  \begin{align*}
   &\mathcal{I}(\theta;[D^q,\mathcal{M}]|D^q)  \\
   ={}&H(\theta|D^q)-H(\theta|D^q,\mathcal{M}) \\ 
   ={}&E[-\log P(\theta|D^q)] + E[\log p([\theta|D^q,\mathcal{M})])] \\
   ={}&E[\log \frac{p(\theta|D^q,\mathcal{M})}{p(\theta|D^q)}] >0 \\
  \end{align*}
\end{proof}

\section{Experimental Details}
\subsection{Personalized Dialogue Generation}
\label{Exp:nlg}

\paragraph{Experimental Setup.}
We implement our model based on the transformer \citep{dehghani2018universal,transformer} with pre-trained Glove embedding \citep{pennington2014glove} following \citep{paml}. The hidden dimensions of the LSTM unit are set to 1024. 
We set the number of neighbors $N = 10$ and the number of local adaptation steps $L = 20$.
We follow all other hyperparameter settings in \citet{paml}: we use SGD for the inner loop training and Adam for the outer loop update with learning rates $0.01$ and $0.0003$, respectively. We set batch size as 16 and use beam search with beam size 5.
\paragraph{Human Evaluation}
\label{human}
We conduct human evaluation following \citet{song2019learning} considering two aspects \textbf{{Quality}} and \textbf{{Consistency}} where five well-educated volunteers annotate 250 generated responses for each model. The annotators score each response from two aspects: \textbf{{Quality}} and \textbf{{Consistency}} in a 3-point scale: 2 for good, 1 for fair, and 0 for bad. \textit{Quality} measures coherence, fluency, and informativeness. 
 \textit{Consistency} measures the task consistency between the generated responses and the person's persona description.
\subsection{Multi-domain Sentiment Classification}
\label{Exp:nlu}

\paragraph{Experimental Setup.}
We utilize a BERT \citep{bert} as the encoder. We fine-tune the off-the-shelf pre-trained BERT on the masked language modeling task following \citep{check} as it greatly improves embeddings' quality \citep{sun2019fine}.
The fine-tuned BERT is then used as the initialization for all few-shot models. 
We use Adam \citep{adam} optimizer for both inner and outer loop update with learning rate $ 2e^{-5} $ and $ 1e^{-5} $ respectively, and we set $\beta=0.2$ in Eqn.~\ref{eq:nlu_prediction}, the number of neighbors $N=20$ and the number of local adaptation steps $ L=5 $. 
\section{Effectiveness of the Interpolation}
\label{Exp:nomm}
To measure whether MemIML improves the learned model initialization, we add an experiment that does not incorporate the memory module during meta-testing (i.e., \ $\beta=1 $ in Eq.~\ref{eq:nlu_prediction}) for the multi-domain sentiment classification task. 
The better result of MemIML than MAML and other regularization methods demonstrate the superiority of our model.
\begin{table}[htbp]
  \centering
  \resizebox{5.2cm}{!}{
\begin{tabular}{cc}
  \Xhline{3\arrayrulewidth}
  Model & Mean Accuracy \\
  \hline   MAML &  82.17  \\
  MR-MAML & 78.14  \\
  Meta-Aug &  83.57  \\
  MetaMix &  83.63  \\
  \hline
  MemIML $ (\beta=1) $ & 84.95 \\
  \Xhline{3\arrayrulewidth}
\end{tabular}}
  \caption{Comparison of mean accuracy on the ARSC.}
\end{table}

\section{Diversity-selection Criterion}
\label{diversity}
For each task-specific memory module $ M $, following \citet{diversity_metric}, we adopt the diversity score as $ S(M)=\mu(M)-\sigma(M) $ on the stored keys, where $\mu(M)=\frac{1}{N^2}\sum _{j=1}^{N}\sum _{h=1}^{N} \angle(K_j,K_h) $ denotes the mean of angles between every two stored key representations and $ \sigma(M)=\frac{1}{N^2}\sum _{j=1}^{N}\sum _{h=1}^{N}(\angle(K_j,K_h)-\mu(M))^{2} $ denotes the variance of those angles \footnote{$\angle(K_j,K_h)=arccos(\frac{K_j \cdot K_h}{\lVert K_j\rVert _2 \lVert K_h\rVert _2})$}.

\ 
\end{document}